\documentclass[conference]{IEEEtran}
\usepackage{etoolbox}
\makeatletter
\patchcmd{\@makecaption}
  {\scshape}
  {}
  {}
  {}
\makeatother
\usepackage{cite}
\usepackage{amsmath,amssymb,amsfonts}
\usepackage{lipsum}
\usepackage{algorithmic}
\usepackage{graphicx}
\usepackage{textcomp}
\usepackage{xcolor}
\usepackage{multirow}
\usepackage{float}
\usepackage{subfig}
\usepackage{multirow}
\usepackage{footnote}
\usepackage{hyperref}
\usepackage{tabularx,booktabs}
\makesavenoteenv{tabular}
\usepackage{threeparttable}

\usepackage[export]{adjustbox}
\usepackage[nottoc]{tocbibind}
\usepackage[ruled,vlined]{algorithm2e}

\def\BibTeX{{\rm B\kern-.05em{\sc i\kern-.025em b}\kern-.08em
    T\kern-.1667em\lower.7ex\hbox{E}\kern-.125emX}}
\begin{document}

\title{BP-Net: Efficient Deep Learning for Continuous Arterial Blood Pressure Estimation using Photoplethysmogram\\

}



\author{
\IEEEauthorblockN{
Rishi Vardhan K\IEEEauthorrefmark{1}, Vedanth S\IEEEauthorrefmark{2}, Poojah G\IEEEauthorrefmark{1}, Abhishek K\IEEEauthorrefmark{1}, Nitish Kumar M\IEEEauthorrefmark{1} and Vineeth Vijayaraghavan\IEEEauthorrefmark{3}
}
\IEEEauthorblockA{
\IEEEauthorrefmark{1}SSN College of Engineering, Chennai, India\\
\IEEEauthorrefmark{2}College of Engineering, Guindy, Chennai, India\\
\IEEEauthorrefmark{3}Solarillion Foundation, Chennai, India\\
\texttt{\{rishivardhan, vineethv\}@ieee.org}\\
\texttt{\{vedanths2000, abhishekk2808, nitish.visva\}@gmail.com}\\
\texttt{poojah17112@ece.ssn.edu.in}
}
}


\maketitle
\begin{abstract}
Blood pressure (BP) is one of the most influential bio-markers for cardiovascular diseases and stroke; therefore, it needs to be regularly monitored to diagnose and prevent any advent of medical complications. Current cuffless approaches to continuous BP monitoring, though non-invasive and unobtrusive, involve explicit feature engineering surrounding fingertip Photoplethysmogram (PPG) signals. To circumvent this, we present an end-to-end deep learning solution, BP-Net, that uses PPG waveform to estimate Systolic BP (SBP), Mean Average Pressure (MAP), and Diastolic BP (DBP) through intermediate continuous Arterial BP (ABP) waveform. Under the terms of the British Hypertension Society (BHS) standard, BP-Net achieves Grade A for DBP and MAP estimation and Grade B for SBP estimation. BP-Net also satisfies Advancement of Medical Instrumentation (AAMI) criteria for DBP and MAP estimation and achieves Mean Absolute Error (MAE) of 5.16 {\em mmHg} and 2.89 {\em mmHg} for SBP and DBP, respectively. Further, we establish the ubiquitous potential of our approach by deploying BP-Net on a Raspberry Pi 4 device and achieve 4.25 {\em ms} inference time for our model to translate the PPG waveform to ABP waveform.
\end{abstract}


\begin{IEEEkeywords}
Blood Pressure, Photoplethysmogram, Arterial Blood Pressure, U-Net, wearable biomedical applications, non-invasive 
\end{IEEEkeywords}

\section{Introduction}

According to the World Health Organization 2019 statistics, Cardiovascular Diseases (CVDs) contribute to nearly 34\% of all deaths worldwide. The most critical risk factor for CVD is elevated blood pressure, also known as hypertension \cite{paper28}. Thereby early diagnosis of abnormal BP can aid a person in acquiring timely treatment and avoid facing severe medical complications by CVDs.

Blood pressure is a vital physiological indicator of a person's heart condition \cite{paper30}. When the heart contracts, BP in blood vessels reaches its maximum value called Systolic Blood Pressure (SBP), and when the heart relaxes, BP in blood vessels reaches its minimum value called Diastolic Blood Pressure (DBP). Additionally, the average BP in a cardiac cycle is termed as Mean Average Pressure (MAP). Hypertension occurs when an individual at rest has SBP more than 140 mmHg or DBP more than 90 mmHg \cite{paper53}. Conventional BP estimation in a clinical setting is performed using a cuff-based Sphygmomanometer that requires the aid of a medical expert. Various factors like mental stress, diet, etc., contribute to fluctuations in BP over time \cite{paper54}. Thereby intermittent estimation of BP by Sphygmomanometer is not reliable for unstable BP measures. The variable nature of BP has necessitated the need for beat-to-beat BP analysis such as Blood Pressure Variability (BPV)\cite{paper36} and continuous BP monitoring. 

The Invasive Arterial Line (IAL) \cite{paper32} approach is considered as the gold standard for continuous BP estimation. The IAL procedure follows the insertion of Intra-arterial catheters in arteries of high-risk or critically ill patients \cite{paper55}. Though known for its superior performance, the method underlies the risk of medical complications such as infection, bleeding, clots, and nerve damage due to its invasive nature. As an alternate to pervasive monitoring, the emergence of cuffless BP estimation methods \cite{paper56} offered a ubiquitous solution that is unobtrusive and non-invasive. PPG signals in the interpretation of various physiological parameters have received widespread attention due to their potential to detect CVDs \cite{paper31}. Cuffless methods projected predominant use of Photoplethysmogram (PPG) signal and its derivatives. 


PPG signal is a low-cost and straightforward representation of the heart's volumetric variation of blood flow. It is measured by an oximeter that illuminates the skin, and the reflection obtained is directly correlated to the changes in the volume of blood flow. The versatility of the PPG signal in terms of inference to efficiency ratio makes it a suitable prospect for estimating blood pressure in a resource-constrained environment. In recent years, optimizing deep learning models for real-time inference on resource-constrained devices has gained prominent interest \cite{paper33}. There is a dearth of work in deep learning-based BP prediction approaches experimented on an edge platform for BP estimation.

This work proposes BP-Net, a signal-to-signal translation U-Net architecture that estimates Arterial BP (ABP) waveform from PPG signal input. Following inference of ABP, we derive SBP and DBP measures and benchmark our results based on international standards. We further experiment with the real-time inference of BP-Net on a resource-constrained edge device and evaluate performance based on inference time. 

The paper is structured as follows, Section II details current work performed under blood pressure prediction using deep learning approaches, Section III comprises dataset and model architecture information. Section IV contains experimental results of BP-Net based on international standards and also discusses how BP-Net compares with existing approaches. In Section V, we conclude the paper with a scope for future work. 

\section{Related Work}

Prior research on blood pressure estimation can be categorized into two groups, Pulse Transit Time (PTT) Technique, and Regression Technique. Pulse Transit Time is the time taken by a blood wave to propagate between two places in a cardiovascular system. PTT is measured as the time interval between the R peak of the Electrocardiogram (ECG) and the systolic peak of fingertip PPG in a cardiac cycle. Since PTT is observed to be negatively correlated with BP \cite{paper35}, different approaches have been proposed to predict BP from PTT by calibration procedures \cite{paper37, paper43, paper44}. 

Several machine learning approaches to BP estimation are based on the Regression Technique. Kachuee \textit{et al}. \cite{paper7} experimented with standard machine learning models like Support Vector Machine, Random Forest to estimate SBP and DBP by feature extraction from PPG and ECG signals. The authors of \cite{paper38} reviewed the problem of accuracy reduction in ML models and proposed Recurrent Neural Network (RNN) architecture with Long Short Term Memory (LSTM) networks for long-term BP prediction. Lee \textit{et al}. \cite{paper39} used a combination of ECG, PPG, and Ballistocardiogram (BCG) signals to train a Bi-LSTM network for beat-to-beat continuous BP estimation. Major prevalent regression-based approaches map input PPG signal and ECG signal or physiological parameters to output SBP and DBP values. Although most of the approaches provide exceptional results, they require extensive feature engineering, ECG signal (obtrusive in measurement), or both. Slapničar \textit{et al}. \cite{paper40} and Shimazaki \textit{et al}. \cite{paper41} experimented with raw PPG signal as input along with its first and second-order derivatives to estimate SBP and DBP. However, performance-wise their approaches did not generalize well compared to existing methods. 

In recent times, similarities between ABP and PPG waveform have attracted considerable interest \cite{paper42}\cite{paper49}. Considering the analogous relationship between ABP and PPG, Ibtehaz \textit{et al}. \cite{paper5} proposed PPG2ABP, a cascaded U-Net architecture to estimate ABP waveform from PPG waveform. From the estimated ABP waveform, DBP and SBP are derived by standard peak detection algorithm \cite{paper43}. Similarly Athaya \textit{et al}. \cite{paper6} performed signal-to-signal translation from PPG to ABP using a U-Net approach and Harfiya \textit{et al}. \cite{paper8} used PPG waveform along with its derivatives to train a LSTM network to estimate ABP. 

Majority of current-day wearable devices that estimate BP utilize the PTT \cite{paper20} approach due to its non-invasive requirements. Since ABP waveform requires minimal pre-processing for estimation and also provides additional diagnostic information about the patient \cite{paper50}, we implement an ABP-based BP estimation framework to be deployed on edge devices that alleviates extensive feature engineering involved with prevailing PTT-based approaches while providing appreciable performance in real-time.

\section{Methodology}

\subsection{Dataset Description}

Physionet's Multi-parameter Intelligent Monitoring in Intensive Care (MIMIC) II Waveform database \cite{paper23} comprises recordings of various physiological signals and physiological parameters from Intensive Care Unit (ICU) patients. For our experimentation, we use MIMIC II derived cuffless Blood Pressure Estimation Data Set compiled by Kachuee \textit{et al}. \cite{paper7}. The dataset contains pre-processed waveform data of ECG, PPG, and ABP signals sampled at 125 Hz. Signals with unusual values of BP such as very high/low (SBP $\geq$ 180, SBP $\leq$ 80, DBP $\geq$ 130, DBP $\leq$ 60) or missing data were excluded from the dataset. Table I presents the statistics of the dataset.

\begin{table}[!ht]
\renewcommand{\arraystretch}{1.3}
\caption{Blood Pressure Ranges in the dataset}
\label{tab:Ranges}
\centering
\begin{tabular}{|l|l|l|l|l|}
\hline
 &
  \begin{tabular}[c]{@{}l@{}}Min\\ (mmHg)\end{tabular} &
  \begin{tabular}[c]{@{}l@{}}Max\\ (mmHg)\end{tabular} &
  \begin{tabular}[c]{@{}l@{}}STD\\ (mmHg)\end{tabular} &
  \begin{tabular}[c]{@{}l@{}}Mean\\ (mmHg)\end{tabular} \\ \hline
DBP & 60.2 & 128.3 & 9.2  & 70.9  \\ \hline
MAP & 68.6 & 136.2 & 9.7  & 93.2  \\ \hline
SBP & 81.5 & 178.8 & 18.7 & 137.9 \\ \hline
\end{tabular}
\end{table}

\subsection{Data Preprocessing}

To remove noise from the raw extracted physiological signals, Kachuee \textit{et al}. \cite{paper7} performed Discrete Wavelet Decomposition (DWT) \cite{paper24} to 10 decomposition levels with Daubechies 8 (db8) as the mother wavelet. Compared to existing filtering methods, the DWT technique is adopted due to better phase response, efficiency in terms of computational complexity, and adaptability to different Signal to Noise Ratio (SNR) regimes. Following DWT, very high-frequency components between 250 Hz and 500 Hz and very low-frequency components corresponding to the range of 0 to 0.25 Hz were eliminated by zeroing their decomposition coefficients. Further conventional wavelet denoising is performed on the remaining decomposition coefficients with soft Rigrsure thresholding \cite{paper25}. Finally, reconstruction of the decomposition is carried out to output a clean processed signal. 

\begin{figure*}
    \centering
   \includegraphics[width=\textwidth, height=8cm]{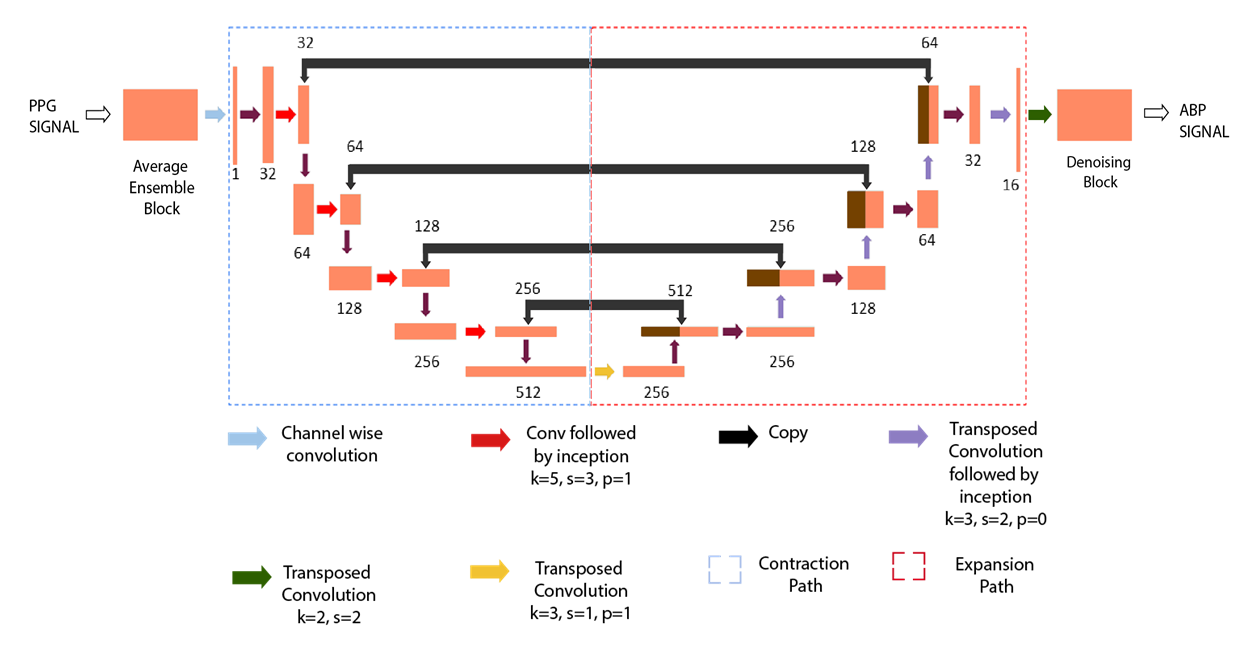}
    \caption{BP-Net architecture}
\end{figure*}

Considering the computational need required to handle the extensive data ($\approx$741.53 hours) obtained, the pre-processed data is subjected to down-sampling, prioritizing the preserving of important information. The down-sampling technique captured 948 subjects worth 127260 counts of episodical data ($\approx$353.5 hours) with each episode attributing to 10-second long waveform data.

\subsection{BP-Net Architecture}

Motivated by the advancements of U-Net in several medical domain applications \cite{paper52, suresh2020endtoend}, BP-Net was developed as an extension of the U-Net framework proposed by Ronneberger \textit{et al}.\cite{paper22}. Analogous to standard encoder-decoder network, the U-Net consists of a contraction path (encoder) and an expansion path (decoder) bridged by skip connections between symmetrical layers. The architecture of BP-Net combines various blocks serving different purposes. The sequence of flow of input is initially through the Average Ensemble Block followed by Contraction Blocks (CB), Expansion Blocks (EB), and ultimately through the Denoising Block. Additionally, the interior of CB and EB are supplemented by the Inception-Residual (IR) Block. The implementation details of BP-Net are described in Section IV A. The design and function of each block are as follows: 

\subsubsection{\textbf{Average Ensemble Block}}

Before forwarding the input signal to the contraction path, the signal is processed to improve the Signal to Noise ratio by subjecting it to Ensemble Averaging. Multiple variants of the input signal are created by passing the signal through a convolutional layer, thereby increasing the number of channels. Further averaging is performed by convolution across the channels to derive a representative signal that is jitter-free. The Ensemble Averaging action leads to faster convergence during training. 

\subsubsection{\textbf{Inception-Residual Block}}

The IR block features the use of multiple convolutional filters of different kernel sizes to perform simultaneous convolutions. Further channel-wise concatenations of the simultaneous convolutions are performed to produce the output. The IR block also contains a residual connection to mitigate the problem of vanishing gradients.

\subsubsection{\textbf{Contraction Block}}

The Contraction Block accomplishes down-sampling operation by subjecting the input through padded convolutional layers to double the number of channels. The output from the padded convolutional layers is further passed onto batch normalization followed by the operation of Leaky ReLU activation. Strided convolution is performed on the activation outputs, and eventually, the intermediate output is passed to the IR block to produce the Contraction Block's output feature map. 

\subsubsection{\textbf{Expansion Block}}

The Expansion Block performs an up-sampling operation by using padded convolutional layers to halve the number of channels. Further, batch normalization and Leaky ReLU activation operations are performed. Strided transposed convolution is carried out on the activation output to reduce the number of channels and pass it to the IR block. Furthermore, for the projection of features from the contraction path to the expansion path, the final EB output is produced by concatenation of the output feature map of the previous EB in the expansion path with the output feature map of the corresponding CB in the contracting path.

\subsubsection{\textbf{Denoising Block}}

The Denoising Block present at the end of the architecture produces the final output of the network by learnt up-sampling to match the ground truth output dimension. It also performs a denoising operation to output a less-distorted signal. 

\subsection{Self Supervised Pretraining}

Unsupervised learning methods for encoder-decoder architecture focus on minimizing the reconstruction error. Although Unsupervised learning leads to successful data representation, it suffers from a significant drawback where the mechanism of model learning depends entirely on single-point model abstraction, i.e., the network learns to construct its output while neglecting other data points present in the dataset. 

Self-Supervised Learning (SSL) aims to understand the semantic relationship between neighboring samples in the dataset to direct learning more representative features and act as a comprehensive feature extraction process. Following the SSL intuition, we initially train the model to reconstruct the input PPG waveform. After training, the learned encoder weights of the model are freezed and eventually used to train another model that performs the required task of reconstructing the ABP signal from the PPG signal. Thereby, the encoder part of the model that captures intermediate waveform representations of PPG signal explicitly, is fine-tuned for learning to estimate the output ABP signal.

\section{Experiments and Results}

\begin{figure}
    \centering
    \subfloat[Ground-Truth]{
        \label{ref_label2}
        \includegraphics[width=0.48\textwidth]{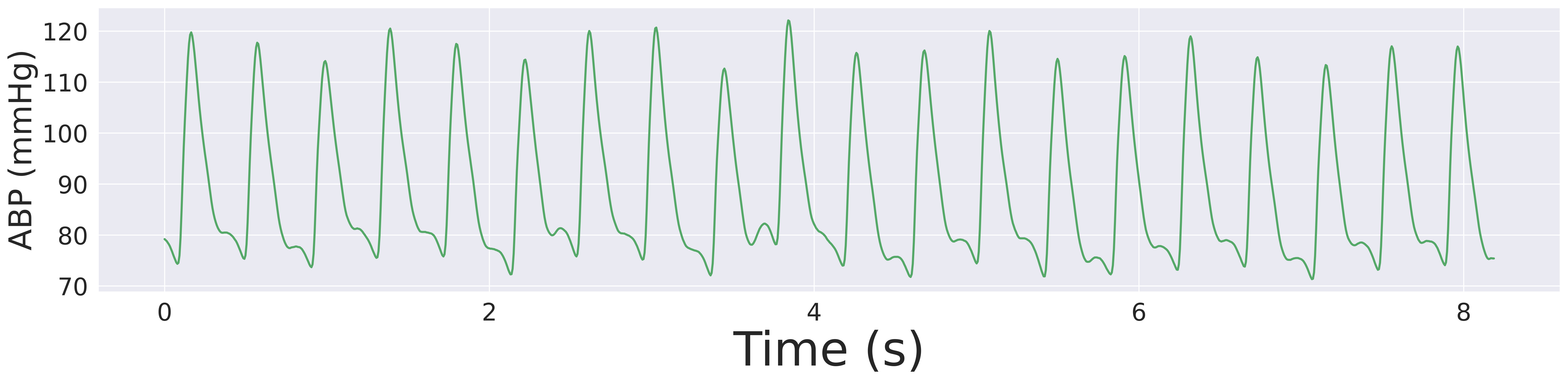}
    }
    \newline
    \subfloat[BP-Net]{
        \label{ref_label2}
        \includegraphics[width=0.48\textwidth]{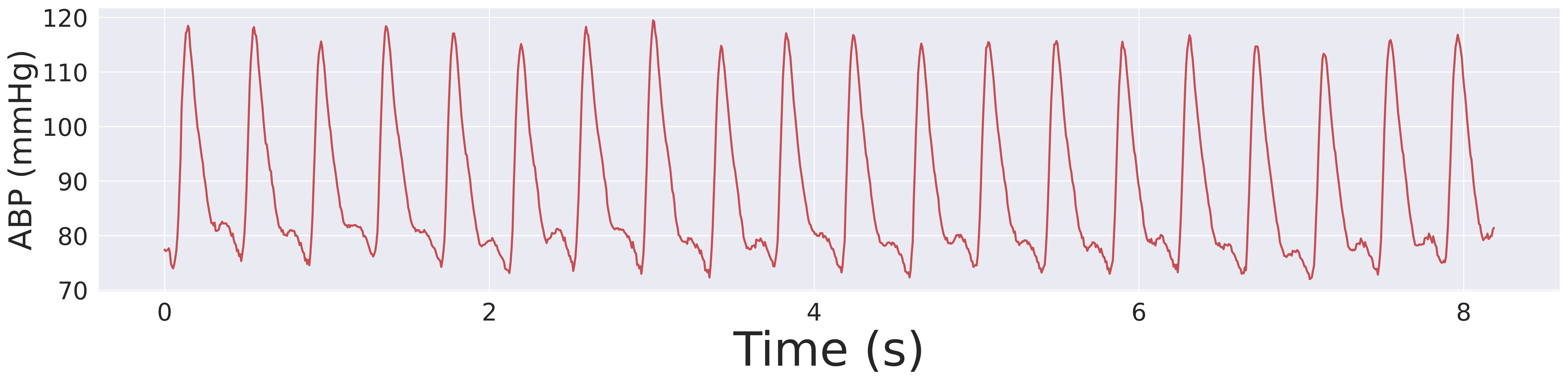}
    }
    \newline
    \subfloat[Comparison]{
        \label{ref_label2}
        \includegraphics[width=0.48\textwidth]{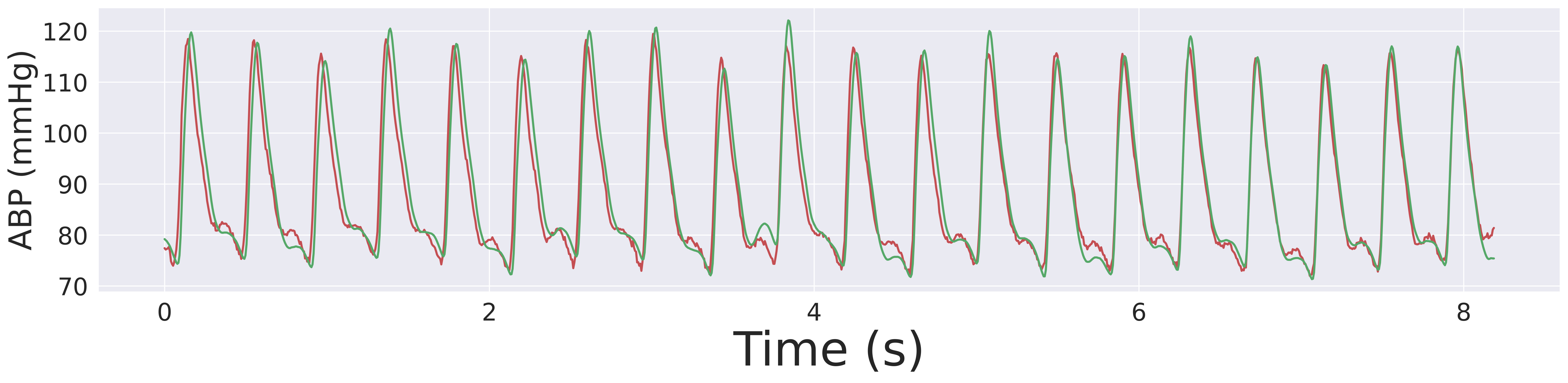}
    }
    \caption{Waveform interpretation of signals}
    \label{ref_label_overall}
\end{figure}

\subsection{Implementation}

For experimentation, the BP-Net architecture from Section III C comprises 5 Contraction Blocks, 5 Expanding Blocks, 1 Average Ensemble block, and 1 Denoising block. Adam optimizer with an initial learning rate of 0.0001 was used to optimize the model's weights to minimize Mean-Absolute-Error (MAE) loss. The hyper-parameters for the model configuration were decided after extensive empirical analysis. 

From the derived 127260 counts of episodic data, 100000 samples were partitioned into training data and 27260 as testing data. The structure of data in MIMIC-II involves every subject's data present next to each other. To reduce the overlap between the training set and testing set, K-Fold cross-validation is suggested \cite{paper7}, and thereby 10-Fold cross-validation is performed for experimentation. Additionally, the input PPG and output ABP signals were mean normalized to facilitate the training need of the deep learning model. 

\vspace{-.5cm}
\begin{align*}
SBP = maximum(ABP) \\
MAP = mean(ABP) \\
DBP = minimum(ABP)
\end{align*}

Considering 10-Fold cross-validation, 10 BP-Net networks were trained each for 300 epochs with a learning rate scheduler that altered the learning rate by a factor of 10 every 100 epochs. The best performing fold was chosen according to the K-Fold cross-validation technique, and the best fold's model is used to evaluate the test data. To derive SBP and DBP values from the predicted values of ABP, the maximum and minimum values of each episode are calculated.

\subsection{Performance Evaluation Metrics}

\subsubsection{\textbf{BHS Standard}}

For structured evaluation criteria to evaluate blood pressure measuring devices and methods, the British Hypertension Society (BHS) \cite{paper26} provides a discrete protocol for evaluation. The BHS standard considers performance accuracy in terms of the percentage of the cumulative error divided across three categories based on performance. For a method to be granted a specific grade, the cumulative error percentages must cross the threshold for a particular grade in every category (5 mmHg, 10 mmHg, 15 mmHg) as detailed in Table II. 

\subsubsection{\textbf{AAMI Standard}}

Similar to BHS, the Advancement of Medical Instrumentation (AAMI) \cite{paper27} also sets rules for validating the effectiveness of the blood pressure measuring devices and methods. According to the AAMI standard, the evaluation criteria are based on whether Mean Error (ME) and Standard Deviation (SD) are within the range of 5 mmHg and 8 mmHg. In addition, the AAMI standard is applicable for evaluation only when a minimum of 85 subjects are involved for BP estimation. 

\subsubsection{\textbf{Mean Absolute Error (MAE)}}

Apart from BHS and AAMI standards, Blood Pressure estimation methods are compared based on Mean Absolute Error. MAE can be formulated as below in Equation 1.
\begin{equation}
MAE = (\frac{1}{N})\sum_{i=1}^{N}\left | e_{i} \right |
\end{equation}

 The {\em e} represents the difference between the ground truth BP and predicted BP value in mmHg, and N represents the number of test samples.

\subsection{Performance Evaluation Results}

\begin{table}[H]
\centering
\renewcommand{\arraystretch}{1.3}
\caption{Evaluation using BHS Standard}
\begin{tabular}{|c|c|c|c|c|}
\hline
\multicolumn{2}{|c|}{\multirow{2}{*}{}} & \multicolumn{3}{c|}{Cumulative Error Percentage}  \\ \cline{3-5}
\multicolumn{2}{|c|}{}                  & \textless{}5mmHg & \textless{}10mmHg & \textless{}15mmHg         \\ \hline
\multirow{3}{*}{BP-Net}    & DBP        & 84.34\%          & 95.19\%           & 98.14\%                \\ \cline{2-5} 
                           & MAP        & 85.64\%          & 94.40\%           & 97.68\%                                \\ \cline{2-5} 
                           & SBP        & 69.21\%          & 86.01\%           & 92.19\%                   \\ \hline
\multirow{3}{*}{BHS}       & Grade A    & 60\%             & 85\%              & 95\%                                        \\ \cline{2-5}
                           & Grade B    & 50\%             & 75\%              & 90\%                                          \\ \cline{2-5}
                           & Grade C    & 40\%             & 65\%              & 85\%         \\ \cline{1-5}
\end{tabular}
\end{table}

\begin{table}[H]
\centering
\renewcommand{\arraystretch}{1.3}
\caption{Evaluation using AAMI Standard}
\begin{tabular}{|c|c|c|c|c|}
\hline
\multicolumn{2}{|c|}{}        & ME            & SD            & Passed \\ \hline
\multirow{3}{*}{BP-Net} & DBP & 0.594         & 4.778         & Yes    \\ \cline{2-5} 
                        & MAP & 0.425         & 4.784         & Yes    \\ \cline{2-5} 
                        & SBP & -0.225        & 8.504         & No     \\ \hline
AAMI Standard           &     & \textless{}=5 & \textless{}=8 &        \\ \hline
\end{tabular}
\end{table}

Table II and Table III present BP-Net's performance based on BHS and AAMI standards, respectively. As observed from Table II, our method yields \textbf{Grade A} for DBP and MAP estimation and \textbf{Grade B} for SBP estimation as per the BHS standard. From Table III, we observe that our method satisfies the requirements for DBP and MAP estimation in the case of the AAMI standard. It is observed that SBP estimation fails in both BHS and AAMI standards by a narrow margin. Subject to the BHS standard, the model falls short by 3\% in the 15 mmHg error threshold while satisfying the 5 mmHg and 10 mmHg error thresholds, thus achieving grade B instead of grade A. While in the case of AAMI, the method fails to satisfy the SD criteria. The inadequacy in the performance of SBP estimation is prevalent in other existent works \cite{paper7, paper5, paper21, paper60} that deal with the MIMIC database. The limitation is generally attributed to the high variance exhibited by the SBP signal (Table 1) compared to DBP and MAP counterparts. Given MAE evaluation, BP-Net achieved MAE values of 5.16 mmHg and 2.89 mmHg for SBP and DBP prediction, respectively.

\subsection{Evaluation of Inference Time}

BP estimation presents a tedious task in terms of continuous BP monitoring. To facilitate the real-time application of our model, inference time can be considered a pivotal evaluation metric. \textit{Inference time} is the time taken by the model to predict real-time input data to produce the desired output. Since our work concentrates on continuous BP monitoring, the time taken by the model to convert PPG to ABP signal is crucial.

\begin{table}[H]
\renewcommand{\arraystretch}{1.3}
\caption{Edge device specification}
\centering
\begin{tabular}{|l|l|}
\hline
SoC             & Broadcom 2711, Quad-core Cortex A72, 64-bit    \\ \hline
RAM             & 4GB                                            \\ \hline
Operating Power & 5V @ 3A                                        \\ \hline
\end{tabular}
\end{table}

To estimate inference time, our model has been deployed on a resource-constrained, low-cost edge device, Raspberry Pi 4 Model B, with the specifications mentioned in Table IV. This resulted in an observed time of 42.53 {\em ms} to convert 10 seconds/1 episode of PPG signal to ABP signal, which translates to 4.25 {\em ms} to convert 1 second of PPG signal to 1 second ABP signal.

Currently, there exists no published work under the context of deep learning-based BP estimation with edge constraints. Thus a general comparison of performances of other works is not possible.

\subsection{Comparison with existing approaches}

A comparative analysis of existing approaches based on MAE and international standards, BHS, and AAMI is presented in Table V. Table V details experimentation results of approaches that map PPG waveform to ABP waveform and successively to SBP and DBP. 

\begin{table}[!ht]

\caption{Results of ABP estimation approaches}
\centering
\renewcommand{\arraystretch}{1.3}
\begin{threeparttable}[ht]
\begin{tabular}{|c|c|c|c|c|c|}
\hline
\multirow{2}{*}{Method} & \multirow{2}{*}{Dataset} & \multicolumn{2}{c|}{MAE} & \multicolumn{2}{c|}{BHS/AAMI\tnote{*}} \\ \cline{3-6} 
 &  & SBP & DBP & SBP & DBP \\ \hline
{\cite{paper6}}   & \begin{tabular}[c]{@{}c@{}}100 subjects\\ (MIMIC II, III)\end{tabular}                & 3.68 & 1.97  & A/P   & A/P  \\ \hline

{\cite{paper5}}   & \begin{tabular}[c]{@{}c@{}}942 subjects\\ (MIMIC II)\end{tabular}                     & 5.73  & 3.45 & B/F   & A/P   \\ \hline
{\cite{paper8}}   & \begin{tabular}[c]{@{}c@{}}5289 subjects\\ (MIMIC II)\end{tabular}                    & 4.05 & 2.41 & A/P   & A/P     \\ \hline

BP-Net & \begin{tabular}[c]{@{}c@{}}942 subjects \\ (MIMIC II)\end{tabular}                    & 5.16 & 2.89 & B/F   & A/P   \\ \hline
\end{tabular}
\begin{tablenotes}
\item[*] BHS, letter represents Grade granted by BHS standard. \\
AAMI, P represents Satisfied and F represents Not Satisfied.
\end{tablenotes}
\end{threeparttable}
\vspace{-2mm}
\end{table}


From the collated information in Table V, Athaya \textit{et al}. \cite{paper6} presents a similar U-Net approach to that of BP-Net to estimate BP. However, they use fewer number of subjects compared to other prominent existing approaches, thereby, cannot be generalized. Harfiya \textit{et al}. \cite{paper8} incorporates first and second-order derivatives along with PPG signal as input to train their model. Although their model achieves exemplar performance for many subjects, the complexity of preprocessing involved makes their approach not feasible for edge deployment. Ibtehaz \textit{et al}. \cite{paper5} makes use of two cascaded U-Net architectures to estimate BP; the computational weight demanded by their approach makes them impractical for inference in a real-time environment. From the perspective of edge implementation, the proposed approach must involve minimal computational power and complexity. BP-Net overcomes the limitations of \cite{paper5, paper8} by using only PPG signal to train a standalone U-Net architecture, thereby reducing the computational complexity involved in porting the model to an edge device. 

Though a diverse amount of work has been performed under blood pressure estimation using deep learning, comparison across the established works remains a difficult task. The main reason for the incongruity is the inconsistent evaluation criteria followed by most of the proposed methodologies. Several works proposed, develop a proprietary dataset of their own and evaluate their work suited to their dataset parameters. This poses an issue considering the number of subjects considered by proprietary datasets tends to be very few compared to public datasets. The requirement of a public dataset is satisfied by the MIMIC database. Although appreciable work is done on MIMIC-II for blood pressure estimation, different works lack a general norm on the number of subjects and the evaluation parameters being used.

\section{Conclusion}
Prevalent non-invasive BP estimation procedures require extensive feature engineering associated with PPG and/or other signals. We alleviate this problem by proposing a Deep Learning based solution to be deployed on resource-constrained devices. In this paper, we develop a U-Net architecture that performs signal-to-signal translation from PPG signal to ABP signal to estimate SBP and DBP values. We further benchmark the inference time of our model on a resource-constrained Raspberry Pi 4 device to validate its application on an embedded or edge platform. Although the performance of BP-Net is comparable to the existing well-known approaches, it can further be improved by increasing the number of subjects taken for experimentation.

\bibliographystyle{IEEEtran}
{\footnotesize \bibliography{citations}}

\end{document}